\documentclass[runningheads]{llncs}
\usepackage{graphicx}
\usepackage{algorithm}
\usepackage{algorithmic}
\usepackage{multicol}
\usepackage{multirow}
\begin{document}
\title{Interpretable Attention Guided Network for Fine-grained Visual Classification\thanks{The work was supported in part by National Natural
Science Foundation of China under Grants 62076016 and 61672079.
This work is supported by Shenzhen Science and Technology Program
KQTD2016112515134654. Baochang Zhang is the correspondence author who is  also with Shenzhen
Academy of Aerospace Technology, Shenzhen, China.}}
\titlerunning{Interpretable Attention Guided Network for FGVC}
%
\author{Zhenhuan Huang\inst{1} \and
Xiaoyue Duan\inst{1} \and
Bo Zhao\inst{1} \and
Jinhu L\"u\inst{1} \and
Baochang Zhang\inst{1}\thanks{Corresponding author}}
\authorrunning{Z. Huang et al.}
%
\institute{Beihang University, Beijing, China \\
\email{\{16231192,17375262,zhaobo0706\}@buaa.edu.cn, jhlu@iss.ac.cn, bczhang@buaa.edu.cn}}
\maketitle              

\begin{abstract}
Fine-grained visual classification (FGVC) is challenging but more critical than  traditional classification tasks. It requires distinguishing different subcategories with the inherently subtle intra-class object variations. Previous works  focus  on enhancing the feature representation ability using multiple granularities and  discriminative regions based on the attention strategy or bounding boxes. However, these  methods highly rely on  deep neural networks   which lack interpretability. We propose an Interpretable Attention Guided Network  (IAGN) for ﬁne-grained visual classification. The contributions of our method include: i) an attention guided framework which can guide the network to extract discriminitive regions in an interpretable way; ii) a progressive training mechanism obtained to distill knowledge stage by stage to fuse features of various granularities; iii) the first  interpretable FGVC method with a  competitive performance on several standard FGVC benchmark datasets.
\keywords{FGVC \and interpretable attention\and knowledge distillation\and progressive training mechanism}
\end{abstract}

\section{Introduction}
Recently, a steady progress has been achieved in generic object recognition with the help of both large-scale annotated datasets and sophisticated model design. However, it is still a challenging task to recognize ﬁne-grained object categories (e.g., bird species \cite{b1}, car models \cite{b2} and aircraft \cite{b3}) which attract extensive research attention. Fine-grained objects are visually similar in global structure by a rough glimpse, while they can be classified into different categories when looking into details, so learning discriminative feature representations from pivotal parts matters in ﬁne-grained image recognition. Existing ﬁne-grained recognition methods can be divided into two groups. One group ﬁrstly locates the discriminative parts of the object and then classiﬁes based on the discriminative regions. Additional bounding box annotations on objects or parts which cost a fortune to collect are commonly required in these two-stage methods \cite{b4,b5,b6}. The other group manages to automatically lead model to focus on discriminative regions via an attention mechanism in an unsupervised manner, which neglects extra annotations. However, these methods \cite{b7,b8,b9,b10} usually need additional network structure (e.g. attention mechanism), and thus generate attention without any interpretability.

\begin{figure*}[htbp]
\centerline{\includegraphics[width=\linewidth]{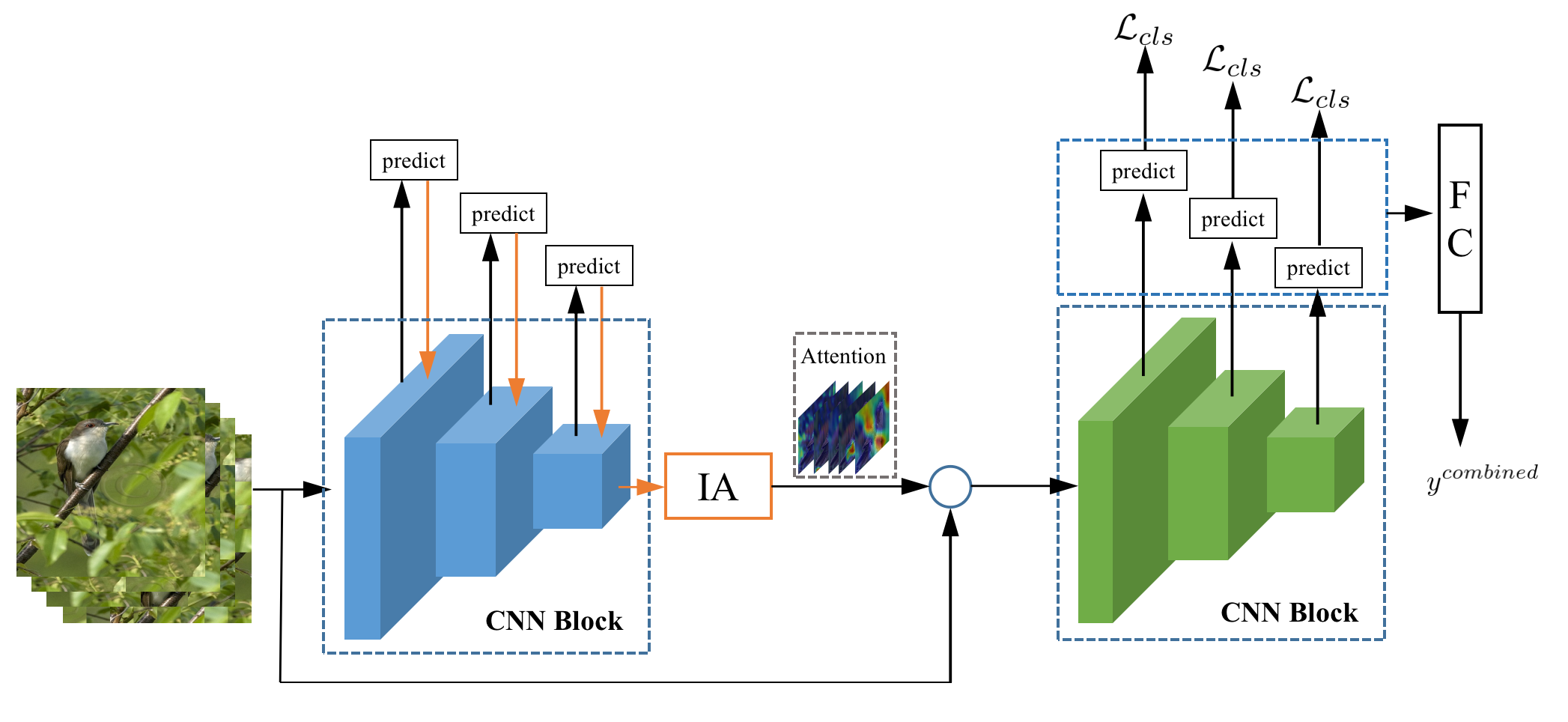}}
\caption{Overview of the interpretable attention guided network. Green and blue blocks denote CNN blocks which share same weights. IA block denotes the interpretable attention generation block. The data that flows along orange arrow represents the process of interpretable attention generation which is not needed in back propagation. Back propagation only flows along black arrows.}
\label{fig1}
\end{figure*}

In this paper, a novel fine-grained image recognition framework named  Interpretable Attention Guided Network  (IAGN) is introduced together with a progressive training mechanism. In addition to the standard classiﬁcation backbone network, we introduce an interpretable attention generation method to automatically learn  discriminative regions, as shown in Fig.~\ref{fig1}. An input image is first carefully augmented by shuffling patches to emphasize discriminative local details. Interpretation attention generation method automatically localizes discriminative regions using in-place back propagation. On the other hand, our data augmentation method and progressive training mechanism further lead our model to recognize from global structure to local details.  Moreover, we also introduce a knowledge distilling mechanism to teach the lower network layers with soft targets generated by a higher network layers, which have a broader receptive field and encode higher-level semantics. Main contributions of this paper can be summarized as follows:
\begin{itemize}
    \item An  Interpretable Attention Guided Network (IAGN)  is introduced for ﬁne-grained visual recognition. It generates attention to lead our model to localize discriminative regions in an  interpretable manner;
    \item  A progressive training mechanism is obtained to distill knowledge stage by stage to fuse features of various granularities; 
    \item Our IAGN  achieves new state-of-the-art or competitive performances on all three standard FGVC benchmark datasets.
\end{itemize}

\section{Related Work}
\subsection{Fine-grained Classification}
Fine-grained image classification methods have been largely improved thanks to the latest development and research findings of convolutional neural networks (CNNs). While some methods attempt to obtain a better visual representation directly from the original image, other techniques try to locate the discriminative regions or parts and learn their features based on the attention generated by the network. Compared with the earlier part/attention based methods, recent research focus has shifted from strongly-supervised learning with annotations of key areas \cite{berg2013poof,zhang2014part,huang2016part}, to weakly-supervised learning with only the supervision of category labels \cite{zheng2017learning,fu2017look,yang2018learning}. 

Recent studies based on weakly-supervised learning mainly address attention to finding the most discriminative parts, more complementary parts and parts of multiple granularities. In order to integrate and fuse information from these discriminative parts better, some fusion methods are put forward. Fu et al. \cite{b7} find that region detection and fine-grained feature learning can promote each other, and thus build a reinforced attention proposal network to obtain discriminative attention regions and multi-scale feature representation based on these regions. Zheng et al. \cite{b8} apply a channel grouping network to jointly learn part proposals and feature representations on each part, and classify these features to predict the categories of the input image. Sun et al. \cite{b10} propose an attention based network, which first apply a one-squeeze multi-excitation module and then put forward a multi-attention multi-class constraint to help to extract multiple region features. Yang et al. \cite{b11} introduce a novel self-supervision mechanism which locates informative regions effectively without bounding boxes and part annotations. 

Inspired by these previous studies, we propose a progressive training mechanism which can distill knowledge stage by stage to fuse features from different granularities and enhance the classification performance. Besides, inspired by the jigsaw puzzle solution, which has been utilized in previous works \cite{cho2010probabilistic,son2014solving,wei2019iterative} and can split the images into pieces to help the network  exploit local regions, we adopt a data augmentation method so that our network would focus more on the discriminative local parts.

\subsection{Interpretable Neural Networks}

Neural Network has achieved huge success in many fields including computer vision, natural language processing and so on these years. However, Neural Network has always been regarded as a “black box” lacking interpretability - we give the network an input, and then get a decision-making result as a feedback, but nobody knows clearly about the decision-making process. Owing to this, it is difficult to convince users of the reliability of Neural Network, resulting in many constraints in its application, especially in security sensitive fields.

The interpretability of neural networks can be divided mainly into two categories: ante-hoc interpretability and post-hoc interpretability. Many recent studies have focused on the latter, which promotes our understanding of neural networks by attempting to interpret trained network models. Zeiler et al. \cite{b12} use deconvolutional networks to visualize what patterns activate each unit. Zhou et al. \cite{b13} utilize global average pooling in CNN to generate Class Activation Maps (CAM), visualizing discriminative regions which CNN draws attention to when classifying the images. Later they further propose a framework called "Network Dissection" \cite{b14}, which quantifies the interpretability of CNN by evaluating the corresponding relationship between a single hidden unit and a series of semantic concepts. 

Our method is based on the Gradient-weighted Class Activation Mapping (Grad-CAM) \cite{b15} method. This technique produces visual explanations for discriminative region decisions of the network, thus making it more interpretable.

\section{Method}

In this section, the proposed Interpretable Attention Guided Network (IAGN) is described. As shown in Fig.~\ref{fig1}, the whole framework of our IAGN includes four parts, which are detailedly described as below.

\subsection{Data Augmentation Method}
\begin{figure}[htbp]
\includegraphics[width=\textwidth]{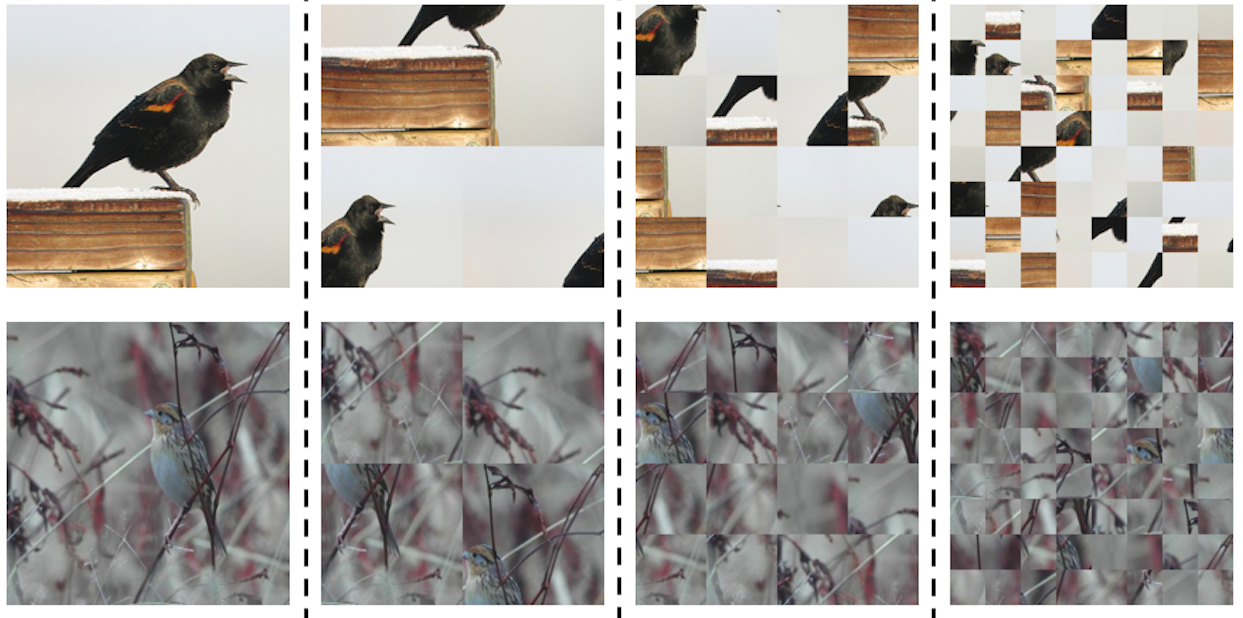}
\caption{Example images for ﬁne-grained recognition and the corresponding shuffled images by data augmentation method.}
\label{fig2}
\end{figure}
In natural language processing \cite{b17}, shufﬂing the order of sequence would help the neural network  find discriminative words while neglecting irrelevant ones. Similarly, in the FGVC task where  local features (more details) instead of global features determine the classification result, shuffling regions of image would promote neural networks to learn from discriminative region details. As shown in Fig.~\ref{fig2}, our data augmentation method is proposed to disrupt the spatial layout of local image regions. Given an input image $I$, we ﬁrst uniformly partition the image into $N \times N$ patches denoted by matrix $R$. $R_{i,j}$ denotes an image patch where $i$ and $j$ are the horizontal and vertical indices respectively ($1\leq i, j \leq N$). In order to destruct global structure but avoid destroying semantics to some extent, patches would be shufﬂed in their 2D neighbourhood. For the $i^{th}$ row of $R$, a new position vector $p_{i}$ of size $N$ is generated, where the $i^{th}$ element $q_{i,j}=i+d$, where $d\sim U(-k, k)$ is a random variable following a uniform distribution in the range of $[-k, k]$. Here, $k$ is a hyperparameter $1\leq k < N$ deﬁning the neighbourhood range. Then we sort the position vector and get a new permutation $\sigma^{row}_{i}$ of patches in $i^{th}$ row subjected to:
\begin{equation}
    \forall j \in {1,2,\dots,N}, |\sigma^{row}_{i}(j) - j| \leq 2k,\label{da1}
\end{equation}
where $\sigma^{row}_{i}(j)$ denotes new vertical index of original image patch $R_{i,j}$.

Similarly, for column $j$, we can get a permutation $\sigma^{col}_{j}$ of patches in $j^{th}$ column subjected to:
\begin{equation}
    \forall i \in {1,2,\dots,N}, |\sigma^{col}_{j}(i) - i| \leq 2k,\label{da2}
\end{equation}
where $\sigma^{col}_{j}(i)$ denotes new horizontal index of original image patch $R_{i,j}$.

Therefore, the original image patch at location $(i, j)$ will be placed at location $(\sigma^{row}_{i}(j), \sigma^{col}_{j}(i))$.
Till now, our data augmentation method has destructed the global structure and ensured that the local region jitters inside its neighbourhood with a tunable hyperparameter. Since the global structure has been destructed, to recognize these randomly shufﬂed images, the classiﬁcation network has to ﬁnd the discriminative regions and learn the tiny differences among categories.

\subsection{Interpretable Attention}
\begin{figure}[htbp]
\centerline{\includegraphics[width=\linewidth]{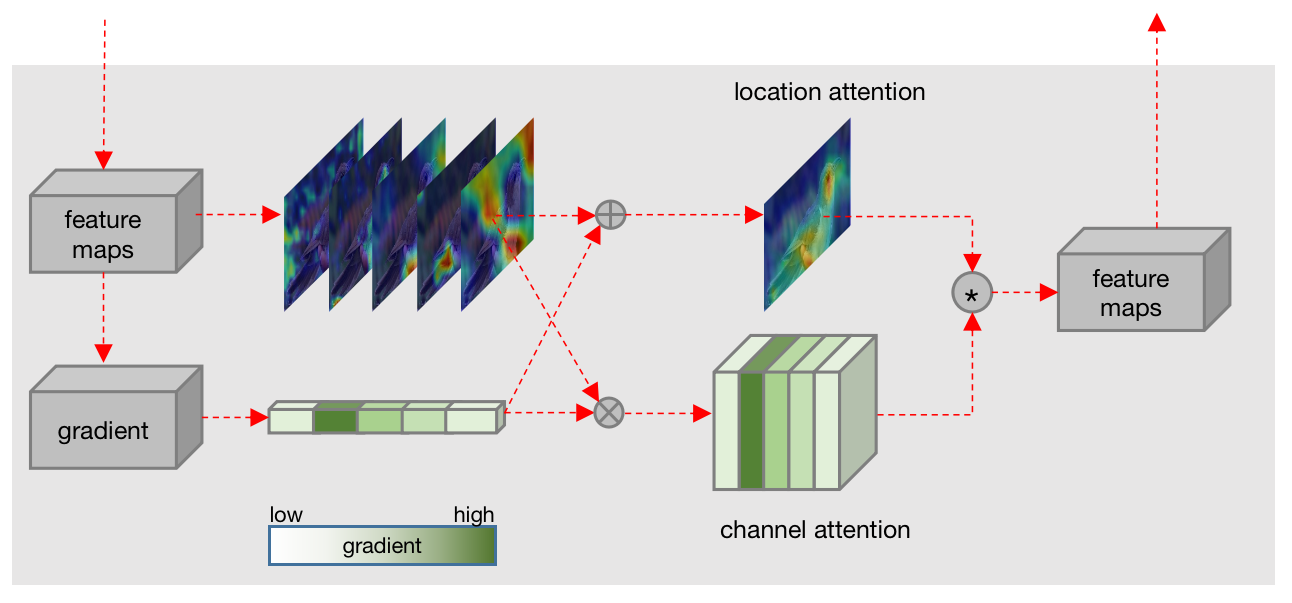}}
\caption{The framework of interpretable attention generation block is composed of location attention and channel attention. $\otimes$ denotes channel wise product operation. $\oplus$ denotes channel wise product and add operation. $\ast$ denotes element wise product.}
\label{fig3}
\end{figure}
To endow the network with the ability to extract discriminative regional features, \cite{b7,b8,b18} crop image or generate attention map via subordinate network for the part localization. To interpret the  localization process, we utilize the Grad-CAM \cite{b15} technique in our network to generate an interpretable attention. 
 
Convolutional layers naturally retain the spatial information which is lost in fully-connected layers. The neurons in these layers extract semantic class-speciﬁc information in the image (object parts). We use the  Grad-CAM technique to obtain the 'importance values' to each neuron in certain layers through back-propogation of the gradient information for a particular decision of interest. Then we will construct our attention map via these 'importance values' which can be interpreted as contribution of each neuron for the final decision. 
As shown in Fig.~\ref{fig3}, in order to obtain the class-discriminative attention map $A^{c}_k \in R^{u \times v}$ of width $u$ and height $v$ for any class $C$, we ﬁrst compute the gradient of the score for class $C$, $y^{c}$ (before the softmax), with respect to feature map activations $F_k$ of the convolutional layer at stage $k$, i.e. $\frac{\partial y^{c}}{\partial F_k}$. These gradients ﬂowing back are global-average-pooled over the width and height dimensions (indexed by $i$ and $j$ respectively) to obtain the channel wise importance weights vector $\alpha ^c_k$:
\begin{equation}
    \alpha ^c_k=\frac{1}{Z}\sum_i\sum_j\frac{\partial y^{c}}{\partial F_k},\label{ia1}
\end{equation}
where $i^{th}$ element of $\alpha ^c_k$ denotes importance value of $i^{th}$ channel in feature map activations $F_k$.

After generating channel-wise importance weights vector $\alpha ^c_k$, we apply it to feature maps and sum in channel dimension to obtain attention map $A_k^c$. 
\begin{equation}
    A^c_k = \sum_{ch=1}^{N_k}softmax(\alpha ^c_k) \odot F_k,\label{ia2}
\end{equation}
where $\odot$ denotes channel wise product, and $N_k$ denotes the number of channels in feature map activations $F_k$.

\subsubsection{Attention Enhancement} When we obtain attention map $A^c_k$ and importance weights vector $\alpha ^c_k$, we apply them to guide the activation propogation. For each channel $F_{k,i} \in F_k$, a new weighted feature map $\hat{F_k}$ is calculated based on attention map and importance weights vector as follows:
\begin{equation}
    \hat{F_k} = (F_k \ast A^c_k) \odot \alpha ^c_k,\label{ia3}
\end{equation}
where $\ast$ denotes element wise product. Through the activation enhancement manipulation, we infuse the interpretable attention information to the feature outputs of convolutional layer, in order to guide the feature learning processing by highlighting the pivotal activations.

\subsection{Progressive Training Mechanism}
As an analogy to recognition process of human, we adopt a progressive training mechanism where we train our network from the higher stage to the lower stage progressively. At the higher stage, the larger receptive ﬁeld and stronger representation ability enable the network to represent high level semantics (say global structure). While at lower stage, receptive ﬁeld and representation ability are limited, the network would be forced to exploit discriminative information from local details (i.e., object textures). Compared to training the whole network directly, this progressive training mechanism allows the model to take a glance at global structure first and then locate discriminative information from local regions for further prediction instead of learning all the granularities simultaneously. 

For the training of the outputs from each stages and the output from the concatenated features, we adopt cross entropy (CE) $\mathcal{L}_{CE}$ between ground truth label $y$ and prediction probability distribution for loss computation as follows:
\begin{equation}
    \mathcal{L}_{CE}(y^l, y)=-\sum_{c}y_c*log(y^l_c),\label{ll}
\end{equation}

and
\begin{equation}
    \mathcal{L}_{CE}(y^{concat}, y)=-\sum_{c}y_c*log(y^{concat}_c).\label{lc}
\end{equation}

As depicted in Algorithm 1, at each iteration, a batch of data $D$ will be used for $S$ steps, and we
only train the output of a certain stage at each step in series except for the first step when we concatenate all outputs. As steps go on, the scale of data augmentation increases. 

\begin{algorithm}
\caption{Progressive Training}
\label{alg:A}
\begin{algorithmic}[1]
\REQUIRE ~~\\
The set of training images for current batch, $D_n$;\\
The set of labels for current batch, $y_n$;\\
Neural network on former batches, $E_n$;\\
Number of progressive training steps, $N$\\
\ENSURE ~~\\
Neural network on current batch, $E_{n+1}$
\FOR{each $i \in [1, N]$}
\STATE {generate shuffled images $D_n'$ using training batch $D_n$};
\STATE compute classification scores $y_i$ for current batch at stage $s_i$;
\STATE{compute cross entropy loss $\mathcal{L}_{CE}^i=-\sum_{c}y_c*log(y^i_c)$}

\STATE compute gradients and update parameters;
\ENDFOR
\STATE $E_n \rightarrow E_{n+1}$
\RETURN $E_{n+1}$

\end{algorithmic}
\end{algorithm}

\textbf{Inference.} During the inference, only the original images will be input into our model and the data augmentation method is unnecessary. In this case, the model combines outputs of the last three stages to obtain $y^{concat}$ for final prediction as shown in Fig.~\ref{fig1}.
For the ensemble purpose, the prediction from various stages contains the information of different granularity, which leads to a better performance when we combine all outputs together with equal weights.

\textbf{Knowledge Distillation.}
Considering our progressive training mechanism, it is natural  to embed knowledge distillation\cite{b16} in our training process. The intention of proposed progressive training strategy is to learn from global structure to local feature for finer recognition. With the help of knowledge distillation, transferring knowledge from stage to stage can be facilitated by a wild margin.


\section{Experiments}
We evaluate the performance of our proposed IAGN on three standard ﬁne-grained object recognition datasets: CUB-200-2011 (CUB) \cite{b1}, Stanford Cars (CAR) \cite{b2} and FGVC-Aircraft (AIR) \cite{b3}. We do not use any bounding box/part annotations in all our experiments. 

\subsection{Implementation Details}
We perform all experiments using PyTorch \cite{b23} with version higher than 1.3 over a cluster of GTX 2080 GPUs. We evaluate our proposed method on the widely used backbone network for classification, ResNet-50 \cite{b24}. This network is pre-trained on ImageNet dataset. The category label of the image is the only annotation used for training. The input images are resized to a ﬁxed size of 512 × 512 and are randomly cropped into 448 × 448. Random rotation and horizontal ﬂip are applied for data augmentation. All above settings are standard in the literature. For all the experiments in this paper, prediction head is plugged into the last three convolutional stages in ResNet-50 backbone, and outputs of these three stages will be concatenated. For training, shuffled scale for data augmentation is set to [1, 2, 4, 8] while concatenated outputs and outputs from stage 5, 4, 3 are used for back propagation respectively at each step. we train our model for up to 150 epochs with batch size as 10, weight decay as 0.0005 and a momentum as 0.9. We use stochastic gradient descent (SGD) optimizer and batch normalization as the regularizer reduced by following the cosine annealing schedule. When testing, image shuffling is disabled, and combined output is the only feature used for location attention generation and final prediction. The input images are center cropped and then fed into the backbone classiﬁcation network for ﬁnal predictions.
\subsection{Performance Comparison}
The results on CUB-200-2011, Stanford Cars, and FGVC-Aircraft are presented in Table~\ref{tab1}. Both the accuracy of $y^{concat}$ and the combined accuracy of all four outputs are listed.

We achieve competitive results on this dataset in a much easier experimental procedure, since only one network is needed during testing. The end-to-end feature encoding methods achieve good performance on birds, while their advantages diminish when dealing with rigid objects. The localization and classification subnets achieve competitive performance on various datasets, usually with a large number of network parameters. For instance, the RA-CNN \cite{b7} consists of three independent VGGNets and two localization sub-networks. By comparison, without extra annotations, our end-to-end approach achieves state-of-the-art and performs consistently well on both rigid and non-rigid objects. Our method outperforms RA-CNN \cite{b7} and MGE-CNN \cite{zhang2019learning} by 3.8\% and 0.6\%, even though they build several different networks to learn information of various granularities, and train the classification of each network separately and then combine their information for testing. This result proves the advantage and validity of our method which exploit multi-granularity information gradually in one network. 

\begin{table*}
\caption{Comparison results on three different standard datasets.}
\begin{center}
\begin{tabular}{|l|c|c|c|c|}
\hline
\multicolumn{1}{|c|}{\multirow{2}{*}{\textbf{Method}}}&\multicolumn{1}{l|}{\multirow{2}{*}{\textbf{Backbone}}} &\multicolumn{3}{c|}{\textbf{Accuracy(\%)}}\\ \cline{3-5} \multicolumn{1}{|c|}{}&\multicolumn{1}{l|}{}&\multicolumn{1}{l|}{\textbf{\textit{CUB-200-2011}}} & \multicolumn{1}{l|}{\textbf{\textit{Stanford Cars}}} &\multicolumn{1}{l|}{\textbf{\textit{FGVC-aircraft}}} \\ \hline\hline
FT ResNet\cite{wang2018learning}& ResNet50& 84.1& 91.7 & 88.5 \\
B-CNN\cite{lin2017bilinear}& ResNet50& 84.1& 91.3& 84.1\\
KP\cite{cui2017kernel}& VGG16& 86.2& 92.4& 86.9\\
RA-CNN\cite{b7}& VGG19& 85.3& 92.5& -\\
MA-CNN\cite{b8}& VGG19& 86.5& 92.8& 89.9\\
MC-Loss\cite{chang2020devil}& ResNet50& 87.3& 93.7& 92.6\\
DCL\cite{b16}& ResNet50& 87.8& 94.5& \textbf{93.0}\\
MGE-CNN\cite{zhang2019learning}& ResNet50& 88.5& 93.9& -\\
S3N\cite{b18}& ResNet50& 88.5& 94.7& 92.8\\ \hline\hline
IAGN& ResNet50& 88.7 &94.0&91.8\\ 
IAGN(combined)& ResNet50& \textbf{89.1}&\textbf{94.8}&92.5\\ \hline
\end{tabular}
\label{tab1}
\end{center}
\end{table*}

\subsection{Visualization}
\begin{figure}[htbp]
\centerline{\includegraphics[width=\textwidth]{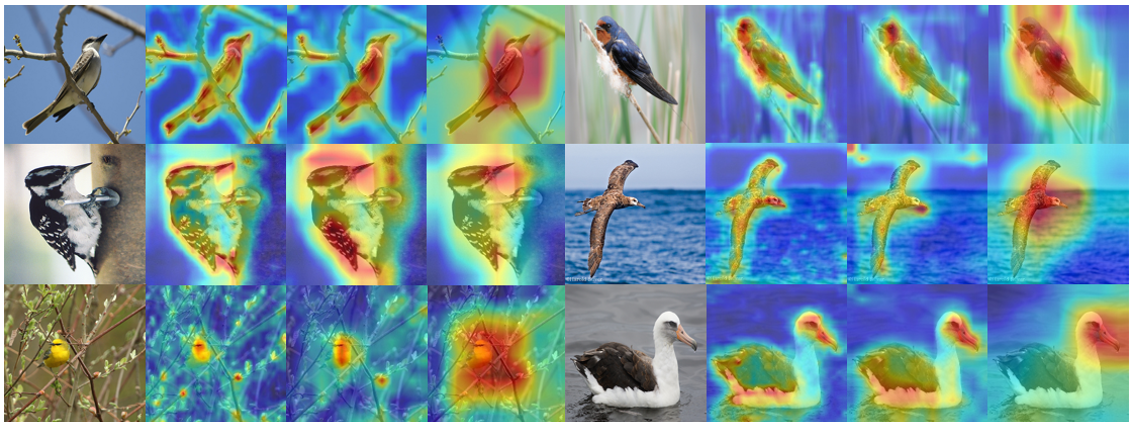}}
\caption{Visualization of interpretable attention generated by the last three stages of our proposed IAGN. The second, third and forth columns corresponds to the attention of the third, the forth and the last stage respectively.}
\label{fig5}
\end{figure}

We visualize the feature maps of the last three convolution layers in Fig.~\ref{fig5}, and we can find that the feature map responses of IAGN are concentrated in discriminative regions. At different stages, the discriminative parts can be consistently highlighted by IAGN model, which demonstrates that our IAGN method is robust. Furthermore, Fig.~\ref{fig5} obviously shows that our model cares about global structure at higher stages and focuses on local details (say discriminative parts) at lower stages. This exactly meets our expectations of the critical advantage of our proposed mechanism.
\section{Conclusion}
In this paper, we propose a novel network named IAGN for fine-grained visual classification. The attention guided framework automatically localizes discriminative regions in an interpretable way. Besides, our data augmentation method and progressive training mechanism further lead the network to implement classification in a global-to-local pattern. Furthermore, knowledge distillation is introduced stage by stage to improve the performance of feature fusion of various granularities. Our method does not require extra regions supervision information and can be trained end-to-end. Extensive experiments against state-of-the-art methods exhibit the superior performances of our method on various ﬁne-grained recognition tasks while maintaining an excellent interpretability.

\bibliographystyle{splncs04}
\bibliography{ref.bib}

%
%
%
%

\end{document}